\documentclass[a4paper]{article}

\usepackage{INTERSPEECH2020}

\title{Stochastic Talking Face Generation Using Latent Distribution Matching}
\name{Ravindra Yadav$^1$, Ashish Sardana$^2$, Vinay P Namboodiri$^3$, Rajesh M Hegde$^4$}
\address{
  $^{1,3,4}$Indian Institute of Technology Kanpur\\
  $^2$NVIDIA}
\email{\{ravin, vinaypn, rhegde\}@iitk.ac.in, asardana@nvidia.com}

\begin{document}

\maketitle
\begin{abstract}
The ability to envisage the visual of a talking face based just on hearing a voice is a unique human capability. There have been a number of works that have solved for this ability recently. We differ from these approaches by enabling a variety of talking face generations based on single audio input. Indeed, just having the ability to generate a single talking face would make a system almost robotic in nature. In contrast, our unsupervised stochastic audio-to-video generation model allows for diverse generations from a single audio input. Particularly, we present an unsupervised stochastic audio-to-video generation model that can capture multiple modes of the video distribution. We ensure that all the diverse generations are plausible. We do so through a principled multi-modal variational autoencoder framework. We demonstrate its efficacy on the challenging LRW and GRID datasets and demonstrate performance better than the baseline, while having the ability to generate multiple diverse lip synchronized videos.
\end{abstract}

\section{Introduction}
Associating a talking face video corresponding to the input of an input speech audio is a challenging problem. More so, as there are a variety of different possible set of generations that could be possible. An ability to solve this problem requires us to have multi-modal distribution association capabilities. To avoid this challenge, current techniques adopt more simplistic one-to-one correspondence. The recent encoder-decoder based approaches enable learning of such multi-modal mappings. We posit in this paper that just learning such one-to-one mappings are highly unsatisfying. Instead, we adopt a principled approach that learns the mapping between the two distributions. This allows us to obtain the samples from the actual distribution of possible generations. We also ensure that these generations are plausible generations.

In order to achieve these aims, we propose a multi-modal generative model for talking face synthesis problem. In this problem, given a single face image of the person and an audio speech segment, the objective is to generate a video clip with that person speaking the given speech. In order to do that, a deep learning model should be able to solve two main challenges: (i) Viseme-to-Phoneme mapping: The problem involves learning the intrinsic correspondence between the facial expressions of the person and the audio segments, both of which possess entirely different statistical properties altogether, and (ii) Diverse predictions: Different speakers can have completely different expressions when speaking the same speech segment, in fact a person can adopt different facial expressions when saying the same speech. Thus the joint distribution between the audio and video modalities is intrinsically a rich distribution that needs to be learnt. This implies that the model should be able to generate multiple different video clips for the same given input image and audio speech.

Recently, various approaches have been proposed to solve talking face generation problem; in these approaches, the focus is on learning a one-to-one mapping function from audio to video domain. Even though these point estimates techniques have shown impressive results, the approaches that can learn rich distributions over output video sequences still remain challenging. In this paper, we propose a stochastic video generation model build upon variational autoencoders. The model is trained with an objective to learn the underlying conditional distribution $p(video|audio)$, and is thus capable of making more diverse predictions that better capture the true variations that are present in real-world data.

Unlike existing approaches, instead of matching the two modalities in the data space, we first map both the modalities into latent spaces and then minimize the distance between the obtained latent distributions. The model thus learns representations that are aligned across both the modalities. These representations are thus robust to type of the modality, such that for any synchronized image and speech segment, we obtain the same latent representation. At test time, we can, therefore, use posterior distribution of source modality (i.e., audio) to generate the corresponding sample of the target modality (i.e., image). In the later sections, we show that using the proposed approach we can the learn the one-to-many mapping between the two modalities, and thus can make multiple diverse predictions.

\section{Related Work}
In this section, we provide relevant background on previous work on audio conditioned video generation, in particular for the "Talking Face Generation" problem. An Audio to Video generation problem is very similar to video generation problems \cite{Srivastava1,Villegas2}. However, the objective now is not only to generate temporally coherent frames but also to have proper synchronization between the given audio and the generated video streams.

In \cite{Suwajanakorn}, an approach for synthesizing President Obama's videos have been proposed. The approach, however, is subject dependent, and thus cannot be used on any arbitrary subjects face images. Also, the model used time-delayed lstm networks to maintain sync between audio and video streams, where the delay hyperparameter is arbitrarily fixed to a particular value.

A simple encoder-decoder based architecture consisting of only Convolution-Deconvolution layers is proposed in \cite{YouSaidThat}. The absence of any recurrent units, however, results in abrupt motions, since the frame generated at any time-step is independent of previously generated frames in the sequence. In addition to encoder-decoder networks, the model also consists of a deblurring module that is separately trained on artificially blurred face images. Thus, the whole network is not end-to-end trainable.

Using GANs, an end-to-end trainable model is proposed in \cite{Vougioukas}. The model consists of a generator network and two discriminator networks (frame and video discriminator). The model uses raw audio as an input for training. A variant of the above model that uses audio Mel Frequency Cepstral Coefficients (MFCC) features is proposed in \cite{Song}; the model consists of an additional discriminator for modeling lips of the speaker in a video. Generative adversarial networks are, however, hard to train and are susceptible to mode-collapse. Our proposed model uses variational autoencoders (VAEs), that are much more stable and provide better likelihoods in comparison to the GAN based networks.

Similar to \cite{YouSaidThat}, the approach followed in \cite{Zhou} generates each frame of the video independent of its previous generated frames. The generated sequence contains random ”zoom-in \& zoom-out” artifacts which thereby necessitate a post-processing stabilization module.

Using attention mechanism, an audio-to-video generation model is proposed in \cite{Hierarchical_pixel_wise_loss}. As in \cite{Ganimation}, an attention mechanism ensures that only pixels that are relevant to audiovisual-correlated regions are modified. The model, however, consists of two networks that are trained separately, hence the model is not end-to-end trainable. Our proposed model, on the other hand, can be trained end-to-end using stochastic gradient descent.

\section{The Proposed Model}
Suppose we are given a video consisting of both audio and frame streams. We represent the frame stream with a set $\mathbb{F}$=\{$f_1$, $f_2$, ..., $f_N$\}, and audio stream with $\mathbb{A}$=\{$a_1$, $a_2$, ..., $a_N$\}, such that both the sets consists of equal number of elements N. In audio-to-video translation, the objective is to learn a mapping function G such that G($a_i$)=$f_i$, where $i$ represent any $i_{th}$ element of the two sets. The function G can be learned as a simple regression function, which simply maps any element $a_i$ in set $\mathbb{A}$ to corresponding element in set $\mathbb{F}$. This, however, could result in an output that have temporal discontinuities. Therefore, to generate any $i_{th}$ frame, we should also take into account the previous generated frames, such that G($a_i$, $f_{i-1}$, $f_{i-2}$, $f_1$)=$f_i$ is an autoregressive function.

\begin{figure}[h]
\centering
\includegraphics[width=5cm]{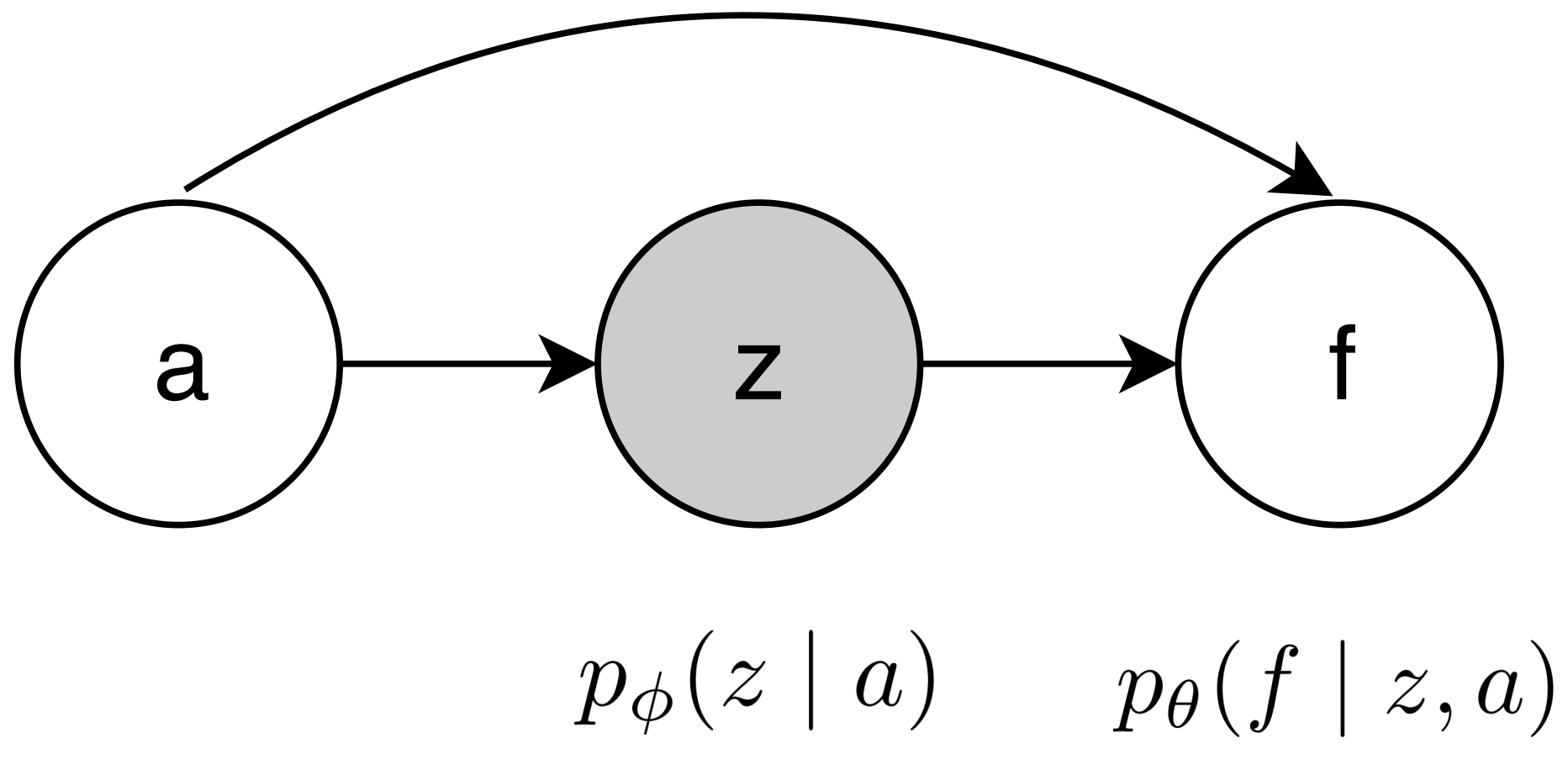}
\caption{\textbf{Conditional Generative model}: Audio sample (indicated as \textit{a}) and frame sample (\textit{f}) are observed variables (white background circles), \textit{z} is a hidden/latent variable (grey background).}
\vspace{-0.1in}
\end{figure}

Since the relationship between the two modalities is intrinsically multimodal, therefore, to force the model to capture all the modes of the conditional distribution $p(\mathbb{F}|\mathbb{A})$, we consider the conditional generative model of the form shown in Figure 1. Given any audio sample \textit{a}, latent variable \textit{z} is sampled from the posterior distribution $p_\phi(z\mid a)$. A frame \textit{f} corresponds to a sample from the conditional distribution $p_\theta(f \mid z,a)$. Thus, the introduction of latent distribution over variable \textit{z} results in learning an one-to-many mapping from audio domain to frame domain. Hence function G can now capture multiple modes of the distribution $p(\mathbb{F}\mid\mathbb{A})$.

\subsection{Model Architecture}
\begin{figure*}[]
\fbox{{\includegraphics[width=\textwidth]{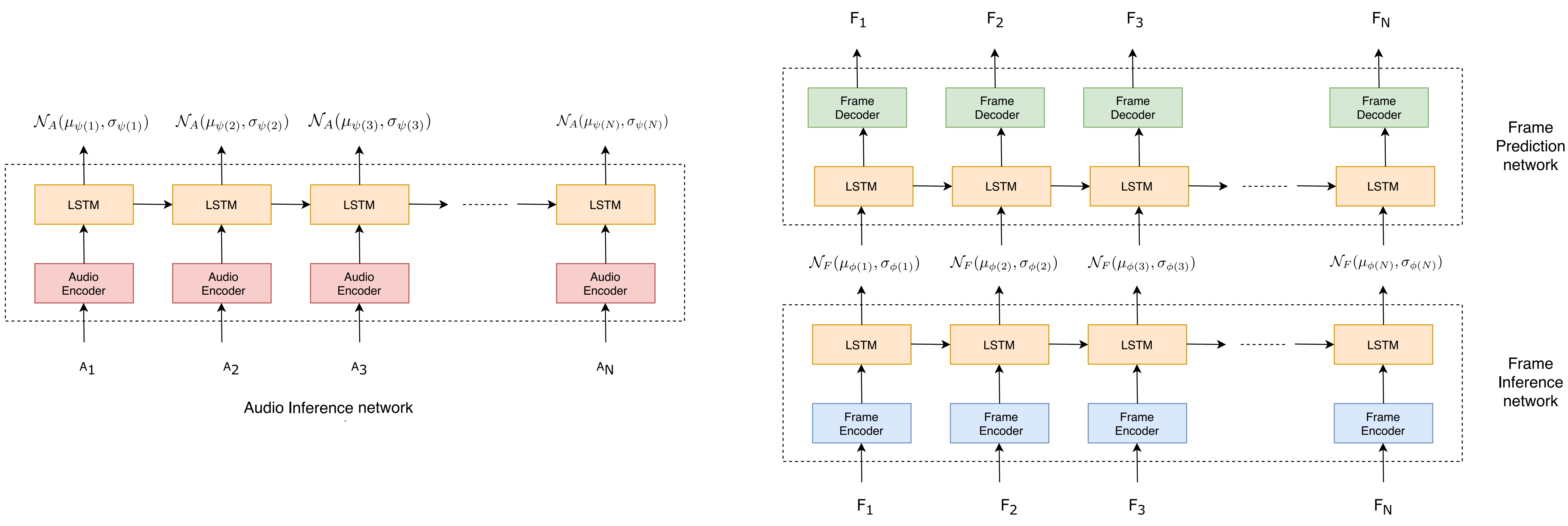}}}
\caption{\textbf{Our proposed model:} The proposed network consists of an audio processing network and a video processing network. At every time step we minimize the KL-divergence between the posterior distributions of the two inference networks, $KL[q_{\phi_{f}}(z|f_t) ||  q_{\phi_{a}}(z|a_t)]$. The frame prediction network is trained to predict the corresponding frame at time t using the sample from the posterior distribution $q_{\phi_{f}}(z|f_t)$. At test time, prediction is based on sample from posterior distribution $q_{\phi_{a}}(z|a_t)$}
\vspace{-0.1in}
\end{figure*}

To learn the generative model of the form $p(\mathbb{F}, \mathbb{A}, \mathcal{Z})$, we adopt the formalism of variational auto-encoders. A variational autoencoder \cite{VAEKingma} is basically a latent variable generative model of the form $p(x)=\int p_\theta(x|z)p(z)dz$, where $p_\theta(z)$ is a prior distribution over the latent variables z (usually assumed to be a standard normal distribution, $p_\theta(z) =  \mathcal{N}(z|0,I)$). The decoder network $p_\theta(x|z)$ is a deep neural network parameterized by $\theta$.

Since the above integral is intractable, therefore, a lower bound on above marginal distribution $p(x)$ (also known as "evidence") is maximized, expressed as,
\begin{equation}
\begin{array}{r@{}l}
\begin{aligned}
\mathbb{E}_{q_{\phi}(z|x)}[\lambda \textup{log}\hspace{0.05cm} p_\theta(x|z)] - \beta KL[q_{\phi}(z|x) ||  p_\theta(z)]
\end{aligned}
\end{array}
\label{eq:1}
\end{equation}
where KL is the Kullback-Leibler divergence. $\lambda$ and $\beta$ \cite{BetaVAE} are training hyperparameters, and $q_{\phi}(z|x)$ is the encoder (or inference) network with parameters $\phi$. The above unimodal network can be trained end-to-end using reparameterization trick applied to latent variable \textit{z}.

Since the audio-to-video translation problem involves multiple modalities, therefore, we need to adapt the above unimodal VAEs formulation to take into account multiple different modalities. We do so by using two different inference networks, one for each modality. At training time, for each time step t, the output of the audio inference module acts as a prior distribution for the frame inference module.

The overall training objective, thus, involves maximizing the following lower bound on the marginal log-likelihood (the “evidence”),
\begin{equation}
\begin{array}{r@{}l}
\begin{aligned}
\hspace{-0.5cm} \textup{log}\hspace{0.05cm} p(f_{1:N})
& \geq \sum\limits_{t=1}^N \mathbb{E}_{q_{\phi_{f}}(z|f_t)}[\lambda\textup{log}\hspace{0.05cm} p_{\theta_{f}}(f_t|z)] \\
& - \beta KL[q_{\phi_{f}}(z|f_t) ||  q_{\phi_{a}}(z|a_t)]
\end{aligned}
\end{array}
\label{eq:2}
\end{equation}

where, both audio and frame posterior distributions are assumed to be Gaussian with diagonal covariance matrix i.e. $q_\phi(z|\cdot) = \mathcal{N}(z|\mu_\phi(\cdot),diag(\sigma_{2}^{\phi}(\cdot)))$. The frame prediction network $p_{\theta_{f}}(f_t|z)$ reconstruct the given input frame using samples from the posterior distribution $p_f(z_t|f_t)$. We set $\lambda$ and $\beta$ equal to values 1 and 1e-6, respectively, in later experiments.

In Figure 2, we show a pictorial representation of our model. The proposed model can broadly be divided into two main modules: an audio inference module and a frame module. The frame module further consists of an inference network and a prediction network. To learn the temporal continuity in the audio and frame streams, both inference networks and the frame prediction network are designed as LSTM \cite{LSTM} cell chains.

The shown image encoder and audio encoder networks first embed the input frame and audio streams into fixed-size feature vectors. These feature vectors then passed as an input to the LSTMs to learn the temporal information present in both the streams. Instead of generating a vector output, these LSTM cells are designed to generate the parameters $(\mu,\sigma)$ of the posterior distribution $q_\phi(z\mid \cdot)$ ($\cdot$ indicates frame f or audio a).

\begin{equation}
\begin{array}{r@{}l}
\begin{aligned}
h_{f,t} = Enc_f(f_t) \rightarrow (\mu_f(t), \sigma_f(t)) = LSTM_f(h_{f,t}) \\
h_{a,t} = Enc_a(a_t) \rightarrow (\mu_a(t), \sigma_a(t)) = LSTM_a(h_{a,t})
\end{aligned}
\end{array}
\label{eq:3}
\end{equation}

Note that, since both inference networks are LSTM chains, therefore, the posterior distribution at any time instant t is dependent on the posterior distributions obtained at previous time steps. Intuitively it has an advantage that since its the distribution over latent variables that encodes all the variations that are present in the frame stream. Therefore, making posterior distributions time-dependent allows the network to produce a smooth transition between different facial expressions.

Additionally, to enable the network to only focus on parts that are most relevant for talking face generation, that is, the lip movement or eyes blink, we add skip connections \cite{u_net} between the frame encoder network at the frame $F_1$ to the frame decoder network at \textit{t} of the frame inference module. Thus the network can simply copy the information that is almost static, while focusing on facial regions that really do change while talking.

At test time, we give the person's face image as input $F_1$ to the frame inference network, along with the audio stream that is fed to the audio inference network. Later at time \textit{t} a frame is generated by first sampling from the posterior distribution $q_{\phi_{a}}(z|a_t)$. The generated sample $z_t$ is then decoded by the frame prediction network to produce the corresponding frame $f_t$. The presence of LSTMs in frame prediction module forces generation at any time step to be dependent on the previous generations. Thus the resulting video is temporally coherent without any zoom in-and-out artifacts that are present in existing models.

\section{Results}
To evaluate our model, we conducted experiments on two real-world datasets that are captured under different environments. The GRID dataset \cite{gridcorpus}, consists of videos of 33 speakers, each speaking 1000 short sentences, captured in a controlled environment. The LRW dataset \cite{lrw}, on the other hand, have videos that are extracted from British broadcast programs, and consists of both frontal and profile views of the speakers. The dataset contains more than half a million videos categorized into different classes based on word labels.

Since the audio and frame modalities have different sampling rates, therefore to align them, we preprocess both before training. For frame stream, we use Dlib library \cite{dlib} to detect and align the face of a speaker in a video frame to a mean face image. For the audio stream, we follow the same preprocessing process as described in \cite{Hierarchical_pixel_wise_loss}. This finally gives us audio and frame streams that have an equal number of elements.

We compare our model against two state-of-the-art models, both qualitatively and quantitatively. As a subjective evaluation, a user study is performed to assess the perceptual quality of the generated videos. Additionally, we demonstrate that our proposed model is capable of making diverse predictions given the same audio and a person's face.

\subsection{Qualitative visualization}
\begin{figure}[h]
\centerline{\fbox{\includegraphics[width=8cm]{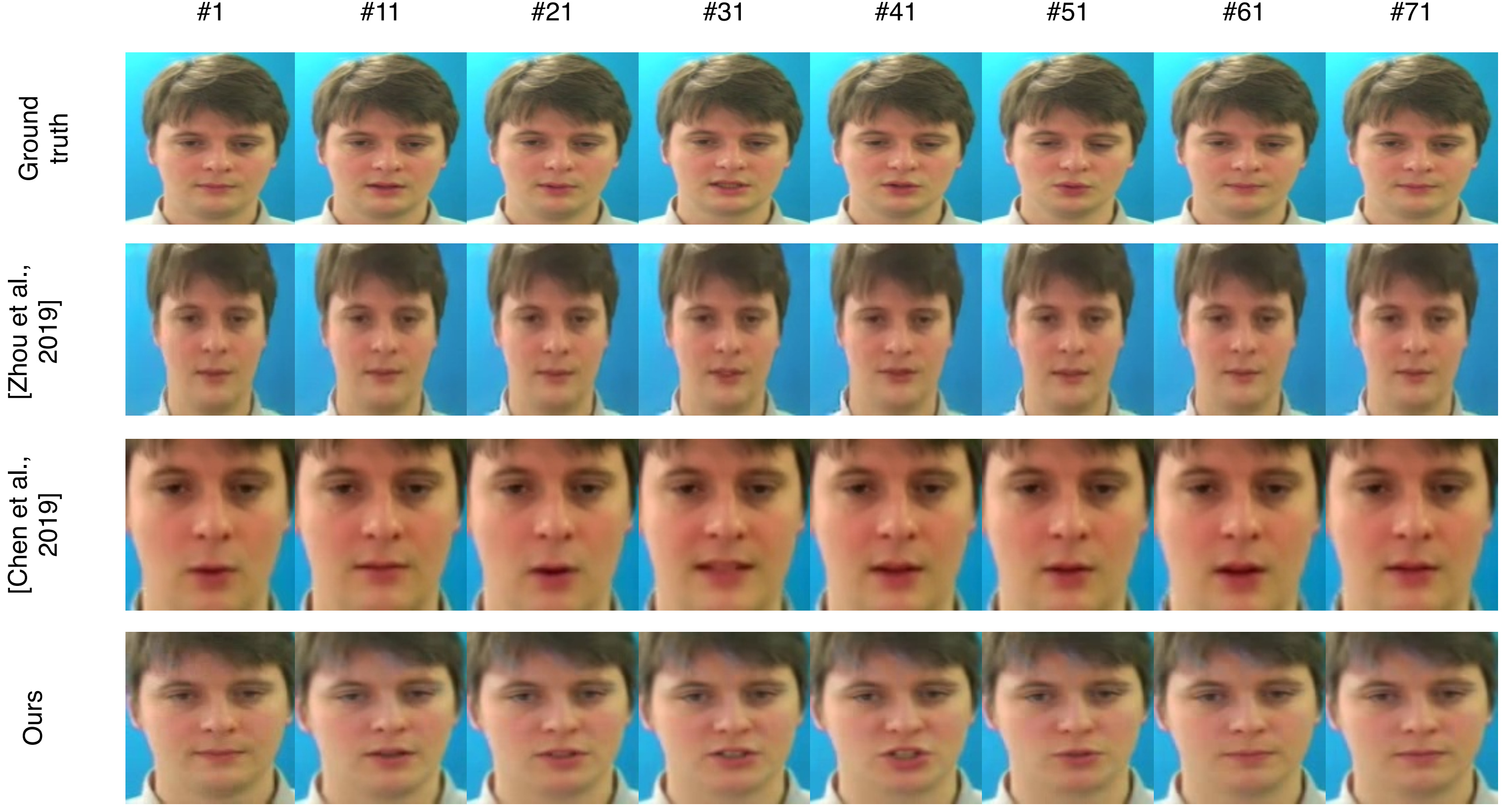}}}
\caption{GRID dataset: Frames generated by different models for the speech input "Bin Blue at C 2 Soon" (\#t indicates the time index)}
\label{fig:gridcorpus}
\end{figure}

\begin{figure}[h]
\centerline{\fbox{\includegraphics[width=8cm]{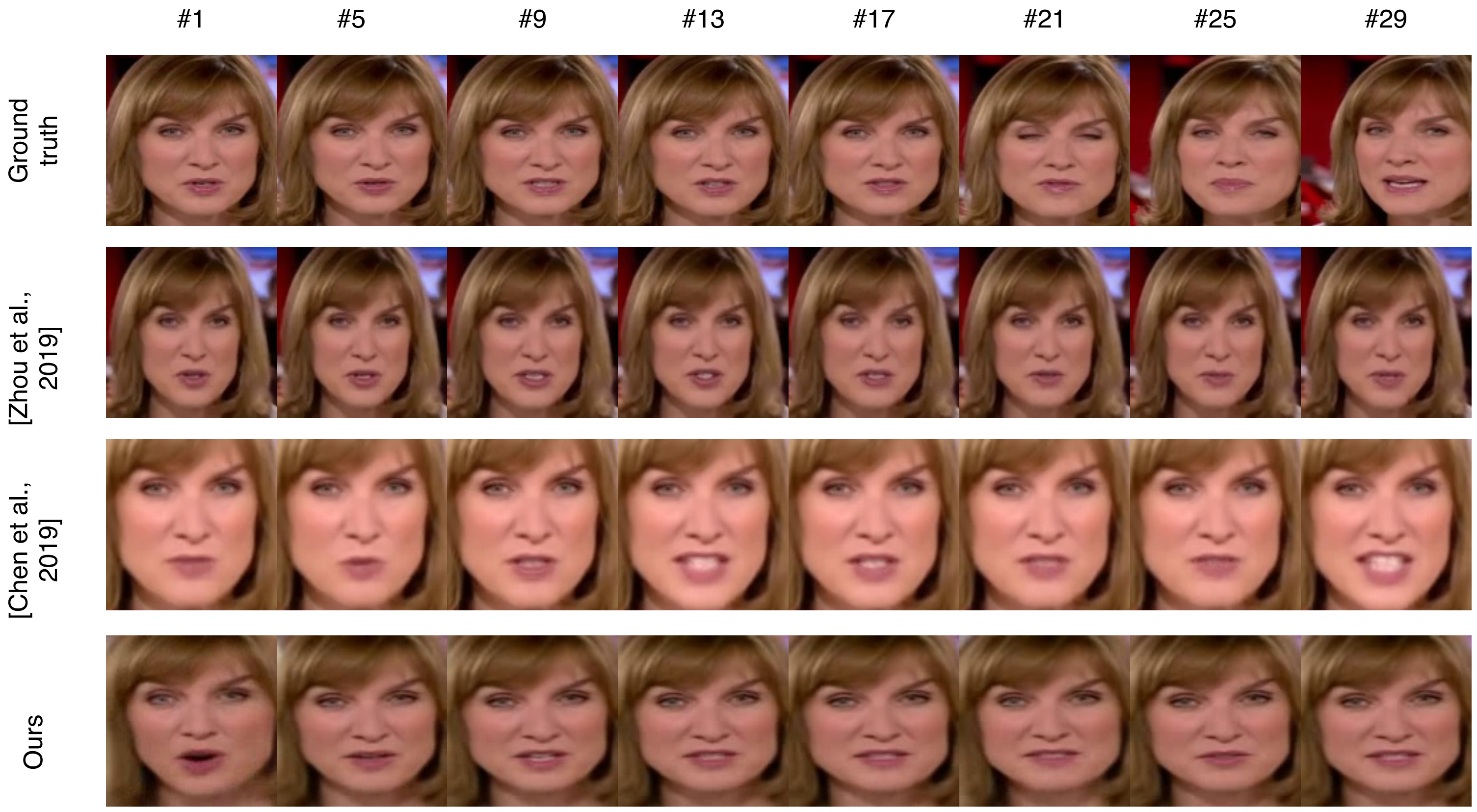}}}
\caption{LRW dataset: Frames generated by different models for the speech input "York is accused of". Notice the similarity between the lip movements in the ground truth and our models generated frames.}
\label{fig:lrw}
\vspace{-0.2in}
\end{figure}

We visualize the generated frames as well as the original video frames in Figure~\ref{fig:gridcorpus} and ~\ref{fig:lrw}. We can see that our model generates lip movements that are very much similar to the ground truth frames. Also, we have observed that our models generated videos look much natural, consisting of movements like eye blinks and head movements (sec 4.3).

The ATVGNet model \cite{Hierarchical_pixel_wise_loss} that directly works on the facial landmarks generates lips movements that look much sharper. But still, theirs end result generated videos look very unrealistic due to the visual analogy approach used for face generation. Also, unlike our unsupervised approach, the model requires annotated landmarks information at training time. The DAVS model \cite{Zhou}, on the other hand, contains random zoom in-and-out artifacts because of the independent generation frames at each time step. For both ATVGNet\footnote[1]{https://github.com/lelechen63/ATVGnet} and DAVS model\footnote[2]{ https://github.com/Hangz-nju-cuhk/Talking-Face-Generation-DAVS}, we use the authors’ publicly available code.

\subsection{Quantitative evaluation}
For quantitative comparison, we evaluate the Structural Similarity Index (SSIM) and Peak signal-to-noise ratio (PSNR) values averaged over randomly chosen 256 test set videos. These values for GRID and LRW dataset are shown in Table~\ref{tab:ssim_psnr}.

\begin{table}[h]
\begin{center}
\caption{Quantitative evaluation: A comparison between average SSIM and PSNR (in dB) values between different models.}
\label{tab:ssim_psnr}
\renewcommand{\arraystretch}{1.2}
\begin{tabular}{|c|c|c|c|c|}
\hline
Model & \multicolumn{2}{c|}{GRID} & \multicolumn{2}{c|}{LRW} \\
\cline{2-5}
 & SSIM & PSNR & SSIM & PSNR \\
\hline
DAVS & 0.82 & 26.95 & 0.80 & 25.10 \\
\hline
ATVGNet & 0.78 & 32.67 & 0.77 & 29.72 \\
\hline
Ours & \textbf{0.86} & 32.01 & \textbf{0.82} & \textbf{30.45} \\
\hline
\end{tabular}
\end{center}
\end{table}

\vspace{-0.2in}
As we can see, our proposed model outperforms or matches the state-of-the-art models in terms of both SSIM and PSNR values. However, as we show in sec 4.3, our proposed model does so while still having the ability to generate diverse results.

\subsection{Diverse predictions}
One of the most crucial aspect of our proposed stochastic framework is its ability to make diverse predictions. Since the space of possible output frames, for a given speech segment, is inherently multimodal. Therefore, the existing models that are based on the simplifying assumption that the future is always deterministic fail to capture the underlying uncertainty. We show that our latent variable model can effectively capture this uncertainty in a principled way.

\begin{figure}[h]
\centerline{\fbox{\includegraphics[width=8cm]{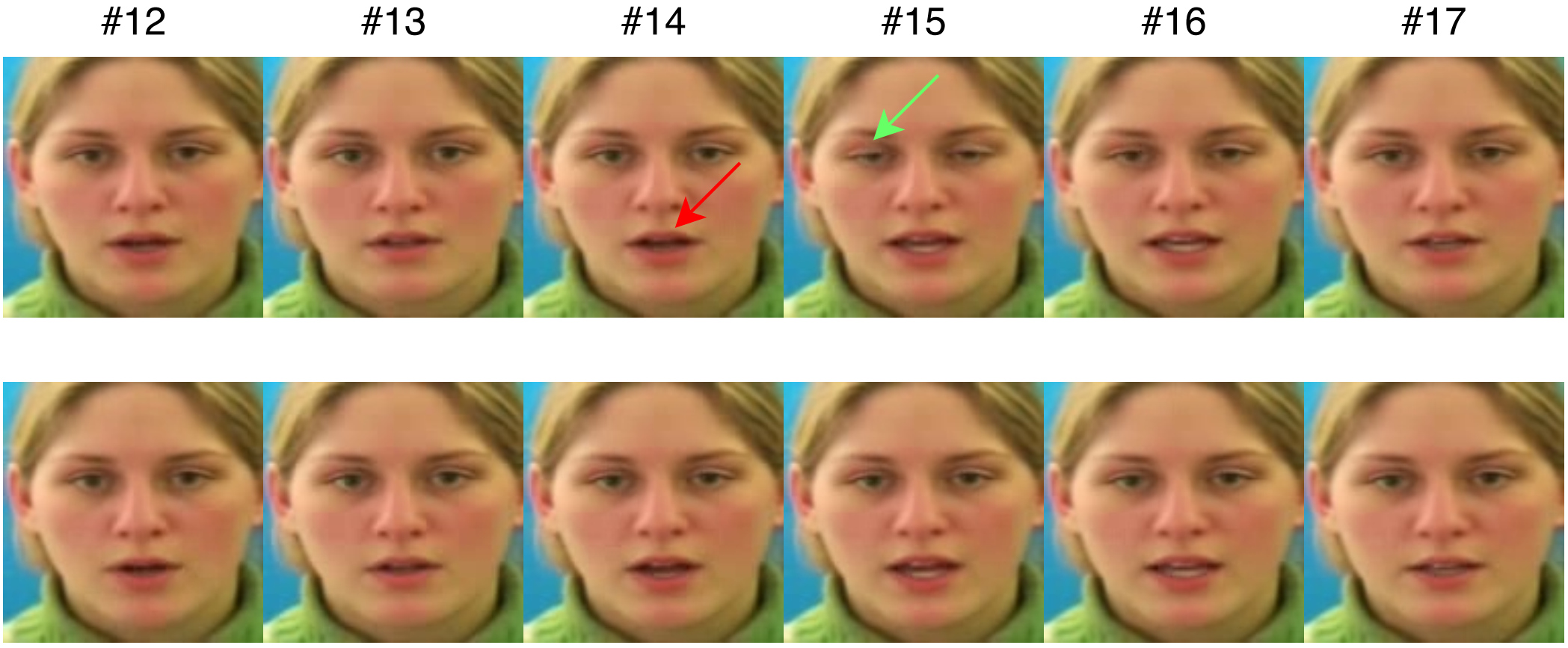}}}
\caption{Diverse predictions: The two rows show frames corresponding to two different output predictions, given the same speech input. We can see the differences, such as the blinking of eyes and lip movements (green and red arrow).}
\label{fig:diversity}
\vspace{-0.2in}
\end{figure}

In Figure~\ref{fig:diversity}, we show two different results produced by our model for the same audio input and face image. We show the differences between the two results by red and green arrows. As we know the subtle movements like eye blinks or different lip movements can make videos look more natural when viewed. The models like ATVGNet fails to capture these small but important nuances, thereby producing videos that look robotic. On the other hand, our latent variable model can generate virtually infinite different videos for different samples of its latent random variables.

\vspace{-0.05in}
\subsection{User Studies}
Even though the two metrics, SSIM and PSNR (sec 4.2), can measure the deviation of the reconstructed frames from the ground truth frames. They cannot measure the subjective aspects that come naturally to human viewers; realism and lip synchronization are two such attributes. Realism defines the naturalness of the videos and can be seen as a measure that depends upon facial expressions like eyebrows and lips movement, eye blinks, etc. Lip sync, on the other hand, measures how well the lip movements of the speaker matches the corresponding audio.

In this study, we showed 75 videos generated by the three models to five participants. These videos were shuffled, so that participants were not aware of which model generated which video. The participants were asked to rate the video on a scale of 1 to 10 (higher score is better). There average responses for the three models are shown in Table~\ref{tab:user_study}. We can see that our probabilistic framework clearly outperforms the two models in terms of both the subjective aspects.

\begin{table}[h]
\begin{center}
\caption{Subjective evaluation: A comparison between Realism and Lip synchronization based on human viewers score.}
\label{tab:user_study}
\renewcommand{\arraystretch}{1.2}
\begin{tabular}{|c|c|c|c|c|}
\hline
Model & \multicolumn{2}{c|}{GRID} & \multicolumn{2}{c|}{LRW} \\
\cline{2-5}
 & Realism & Lip Sync & Realism & Lip Sync \\
\hline
DAVS & 5.14 & 5.2 & 5.92 & \textbf{6.4} \\
\hline
ATVGNet & 5.74 & 5.58 & 5.9 & 6.19 \\
\hline
Ours & \textbf{7.28} & \textbf{7.11} & \textbf{6.68} & \textbf{6.4} \\
\hline
\end{tabular}
\end{center}
\vspace{-0.1in}
\end{table}

\vspace{-0.2in}
\section{Conclusion}
In this work, we presented a generative model for audio-to-video translation, in particular, for talking face generation problem. The proposed stochastic model can capture the underlying uncertainty in future frame prediction in a principled way. To the best of our knowledge, our latent variable model is the first that focuses on the diversity aspect. The fact that instead of point estimates we are learning mappings between distributions also enriches our model. Further, we ensure that the coherence between the speech signal and the mouth movements is preserved. The introduction of latent variables into the modeling process allows us to make multiple predictions given the same input. Our experiments show that our model outperforms the existing models, both qualitatively and quantitatively. But most importantly, the ability to generate virtually infinite results based on the same inputs is the main contribution of this work.

\bibliographystyle{IEEEtran}
\bibliography{mybib}

\end{document}